\documentclass{ecai} 

\usepackage{latexsym}
\usepackage{amssymb}
\usepackage{amsmath}
\usepackage{amsthm}
\usepackage{booktabs}
\usepackage{enumitem}
\usepackage{graphicx}
\usepackage{color}

\usepackage[T1]{fontenc}
\usepackage[misc]{ifsym}

\usepackage{times}
\usepackage{soul}
\usepackage{url}
\usepackage{xcolor}
\usepackage[hidelinks]{hyperref}
\usepackage[utf8]{inputenc}
\usepackage[small]{caption}
\usepackage{amsfonts}
\usepackage{subcaption}
\usepackage{multirow}
\usepackage{arydshln}
\usepackage{wrapfig} 
\newcommand{\mbf}[1]{\mathbf{#1}}
\newcommand{\mbb}[1]{\mathbb{#1}}
\newcommand{\sm}[1]{SMART\textsubscript{#1}}
\newcommand{\tc}[2][blue]{\textcolor{#1}{\textit{#2}}}

\usepackage{amsmath}
\usepackage{centernot}

\newcommand{\BibTeX}{B\kern-.05em{\sc i\kern-.025em b}\kern-.08em\TeX}

\begin{document}


\begin{frontmatter}


\paperid{123} 


\title{SMART: Relation-Aware Learning of Geometric Representations for Knowledge Graphs}

\author[1,2,6]{\fnms{Kossi}~\snm{Amouzouvi}
 \thanks{Corresponding Author. Email: kossi.amouzouviqtu-dresden.de}
 }
\author[3]{\fnms{Bowen}~\snm{Song}
}
\author[4]{\fnms{Andrea}~\snm{Coletta}\thanks{The opinions expressed in this paper are personal and should not be attributed to Banca d'Italia.}} 
\author[4,$**$]{\fnms{Luigi}~\snm{Bellomarini}} 	
\author[5]{\fnms{Jens}~\snm{Lehmann}\thanks{Work done outside of Amazon}} 
\author[1,6,7]{\fnms{Sahar}~\snm{Vahdati}}

\address[1]{ScaDS.AI Dresden/Leipzig, TU Dresden, Germany} 
\address[2]{Department of Mathematics, KNUST, Ghana}
\address[3]{Chinese University of Geosciences, Wuhan, China}
\address[4]{Banca d'Italia, Italy}
\address[5]{Amazon, Dresden, Germany}
\address[6]{InfAI, Dresden, Germany}
\address[7]{TIB
, Hannover, Germany}


\begin{abstract}
Knowledge graph representation learning approaches provide a mapping between symbolic knowledge in the form of triples in a knowledge graph (KG) and their feature vectors. Knowledge graph embedding (KGE) models often represent relations in a KG as geometric transformations. Most state-of-the-art (SOTA) KGE models are derived from elementary geometric transformations (EGTs), such as translation, scaling, rotation, and reflection, or their combinations. 
These geometric transformations enable the models to effectively preserve specific structural and relational patterns of the KG. However, the current use of EGTs by KGEs remains insufficient without considering relation-specific transformations. Although recent models attempted to address this problem by ensembling SOTA baseline models in different ways, only a single or composite version of geometric transformations are used by such baselines to represent all the relations. 
In this paper, we propose a framework that evaluates how well each relation fits with different geometric transformations. Based on this ranking, the model can: (1) assign the best-matching transformation to each relation,  or (2) use majority voting to choose one transformation type to apply across all relations.
That is, the model learns a single relation-specific EGT in low dimensional vector space through an attention mechanism. Furthermore, we use the correlation between relations and EGTs, which are learned in a low dimension, for relation embeddings in a high dimensional vector space. 
The effectiveness of our models is demonstrated through comprehensive evaluations on three benchmark KGs as well as a real-world financial KG, witnessing a performance comparable to leading models. 
\end{abstract}

\end{frontmatter}


\section{Introduction}

Knowledge Graph Embedding (KGE) models are representation learning models that enable the use of machine learning on symbolic representations of domains of interest through \textit{Knowledge Graphs} (KGs).
Intuitively, KGEs serve downstream machine learning tasks on KGs, such as link prediction, question answering, recommendation services~\cite{ji2020survey}, and others. 

\smallskip
\noindent\textbf{Knowledge graph embeddings}. A KGE model takes as input a KG and maps its facts into a feature space. The facts of a KG are typically represented as \emph{true} triples in the form of  $(\mathit{head}, \mathit{relation},\mathit{tail})$, where head ($h$) and tail ($t$) refer to entities linked by a relation ($r$). Through its facts, a KG typically ``talks'', i.e., makes assertions, about real-world entities like ``\textit{empathy is a hypernym of sympathy}''---as we would find in the WordNet knowledge grap~\cite{miller1995wordnet}---or ``\textit{John Doe owns SmartKG Ltd.\ company''}---as it would be the case in a financial KG.


\smallskip
The working mechanism of most KGEs is quite intuitive: they learn a transformation that maps the embedding of the head entity in such a way that, when combined with its relation, it approximates the embedding of the tail entity. In the training, the model optimizes this transformation, by minimizing a loss function that sustains lower scores for \textit{true} (plausible) triples and higher scores for \textit{negative} ones.

In many models, a single \textit{elementary geometric transformation} (EGTs) is adopted. For example, TransE~\cite{bordes2013transe} and RotatE~\cite{sun2019rotate} use a translation or a rotation, respectively; in some cases scaling functions or reflections are also considered. 

\smallskip\noindent\textbf{The problem of complex relations}.
Unfortunately, in complex real-world KGs, a single transformation may fall short of capturing relations with relatively common characteristics, such as symmetry (not supported by simple translations), or absence of commutativity (not supported by rotations). A few more advanced KGE models, such as  QuatE~\cite{zhang2019quaternion}, can support multiple EGTs, yet they do not really combine them, but use the same transformation (one of the mentioned one, for example) to represent all the relations of each input KG. This is still insufficient to harness complex patterns, both reflexive and symmetric, or hierarchical, which cannot be represented by either a single EGT or a fixed combination of them.

\begin{figure*}[h]
    \centering
    \includegraphics[width=\linewidth]{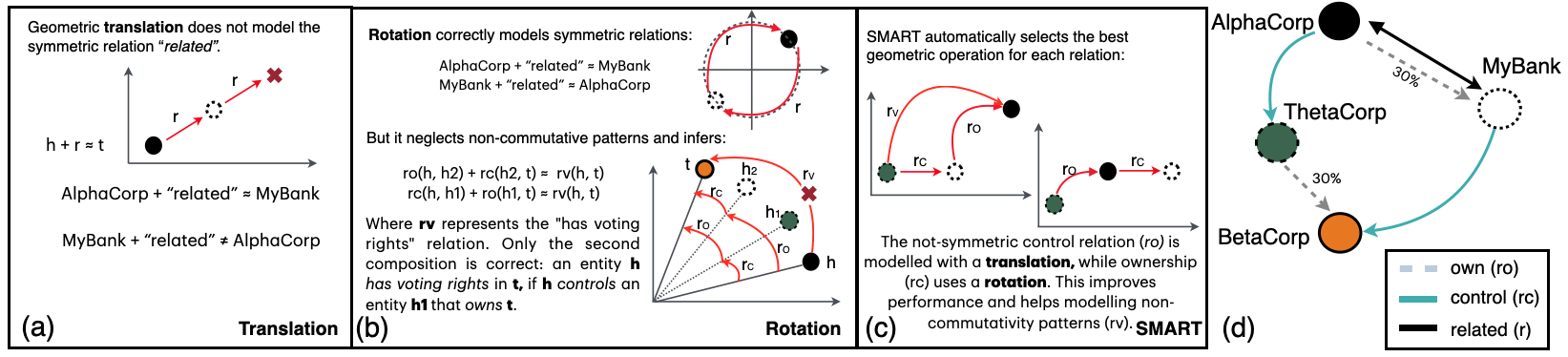}\smallskip
    \caption{An example showing geometric transformation embeddings applied to an anonymized fragment of a company-ownership KG (d). In example (a), a fixed \textit{translation} transformation fails to capture symmetric relations: ``MyBank'' is not correctly \textit{related} to ``AlphaCorp''. In example (b), a \textit{rotation} successfully models such symmetric relation; however it introduces a commutative property, which does not generally hold in these financial structures: \textit{control} + \textit{own} $\implies$ \textit{voting rights}, but  \textit{own} + \textit{control} $\centernot\implies$ \textit{voting rights}. Finally, our approach (c) selects a specific transformation for each relation to correctly model the KG.} 
    \label{fig:example}
    \vspace{0.1in}
\end{figure*}

\smallskip\noindent\textbf{A real-world example}.
Consider the case of the KG of a large central bank in Europe, whose goal is to model the financial relations of European companies. In fact, such technology together with machine learning models would effectively support analysts in answering data-intensive queries and supporting supervision asks~\cite{atzeni2020weaving}. Yet, KGE models relying on a single transformation may struggle to capture the complex and heterogeneous relations typical of the financial domain. For instance,
as shown in Figure~\ref{fig:example}(a): while ``MyBank'' (t) is correctly associated to AlphaCorp (h) using the \textit{related} (r) relation ($h+r\approx t$), the symmetric case is not properly modeled by TransE.
Similarly, while RotatE can effectively model
relational patterns such as symmetry (e.g., \textit{related}) and inversion patterns (e.g., \textit{own} and \textit{is owned}), it introduces a commutative property that does not generally hold in real ownership structures. For instance in Figure~\ref{fig:example}, ``AlphaCorp'' controls ``ThetaCorp'', which in turn owns ``BetaCorp'', leading to a valid inference of voting rights from ``AlphaCorp'' to ``BetaCorp''. Yet, the inverse does not apply: ``AlphaCorp'' owns ``MyBank'', which controls ``BetaCorp'', but it does not necessarily imply that ``AlphaCorp'' holds voting rights in ``BetaCorp''.

\smallskip\noindent\textbf{Contribution}.
Based on this intuition, our research hypothesis is that allowing relation-specific GTs can improve the performance of KGE models. 
In this paper, we propose SMART, a KGE framework designed to learn relation-specific geometric transformation and so overcome the mentioned shortcomings. Rather than fixing a predefined transformation for all relations, SMART selects the most appropriate GT for each relation by ranking multiple candidate transformations, based on how well they model the structure and semantics of each relation. This allows the framework to flexibly adapt to the diverse patterns of a real-world KG. 
%
The EGTs included in our framework are translation, rotation, reflection, and scaling. 

Each relation can be captured by one or many of those GTs, with a three-step approach.
%
In the \textit{training} phase (1), a relation is initially encoded through all the GTs, which contribute equally. 
In the \textit{adaptive learning} phase (2), we dynamically update the attention weights over the set of candidate GTs for each relation, allowing it to gradually focus on the transformation that best capture the relational patterns observed during phase~1.
Finally, during the \textit{freezing} phase (3), the framework selects a subset of the most relevant GTs for each relation, based on the learned attention distribution. Then it continues the training using only the selected transformations, allowing for a more focused and efficient refinement of relation-specific transformations.

\smallskip
Coming back to our financial example, SMART proposes a dedicated EGT for 
each type of financial relation, as shown in Figure~\ref{fig:example}(c): our framework chooses the most appropriate transformation based on the nature of each relation, improving the performance on downstream tasks   
and solving limitations of individual transformations.

\smallskip
In particular, in this paper, we contribute as follows:

\begin{itemize}\setlength\itemindent{0pt}
\item We present \textbf{a general, fully-developed framework for multi-stage and relation-specific optimization of GTs.}
\item We provide a \textbf{theoretical analysis of how the framework captures key relational patterns}, including (anti)symmetry, inversion, composition, and (non-)commutativity.
\item In addition to performance evaluation, we offer an \textbf{empirical analysis of which transformations are most appropriate} for different relations.
\item An open-source \textbf{PyTorch implementation}~\cite{clfgithub} of our framework.
\end{itemize}


\smallskip\noindent\textbf{Overview}. 
In Section~\ref{sec:related}, we review the related work on transformations and KGEs. Section~\ref{sec:preliminaries} provides the necessary preliminaries. In Section~\ref{sec:smart_kge}, we introduce SMART, presenting its formal definition and detailed description. Section~\ref{sec:exp} reports the experiments and evaluation results. Section~\ref{sec:conclusions} concludes the paper. 

\section{Related Work}\label{sec:related}
\paragraph{\textbf{Transformation-based KGE models.}}
Numerous KGE models apply geometric transformations to entities, often without considering special characteristics of relations. 
Foundational models such as such as TransE~\cite{bordes2013transe} and its variants~\cite{ji2015knowledge,lin2015learning,wang2014knowledge} use translation as the main transformation applied to the head entity to predict the tail.
Several other models such as RotatE~\cite{sun2019rotate} rely on rotation (via Hadamard product) in the design of their score functions. 
These models optimize embedding vectors to maximize triple plausibility, typically measured by the distance between the transformed head and tail vectors. 
Additionally, some embedding models assess the plausibility of triples based on the angle of transformed head and tail, such as DisMult~\cite{yang2014embeddingDistmult}, ComplEx~\cite{complex2016trouillon}, QuatE~\cite{zhang2019quaternion}, and RESCAL~\cite{nickel2011three}.

\paragraph{\textbf{Relation-aware transformation models.}} 
More recent efforts have introduced models that account for the unique characteristics of relations ~\cite{weber2018curvature}.  
For example, models such as MuRP, RotH, RefH, and ATTH \cite{balazevic2019multi,chami2020low} focus on hierarchical structures~\cite{suzuki2018riemannian,ji2016knowledge}.  
The Lorentz model~\cite{nickel2018learning} employs a relation-specific embedding, and aims at learning hierarchical patterns from unstructured similarity scores.
MuRP~\cite{balazevic2019multi}, in particular, maps multi-relational KGs on a Poincaré ball~\cite{ji2016knowledge} with a focus on multiple simultaneous hierarchies. 
These mentioned SOTAs have their variants offering alternative formulations such as MuRE, RotE, RefE, and ATTE in Euclidean spaces. 
In the aforementioned approaches, all the models are distance based models, except ComplEx. 
While TransE, RotatE, and ComplEx use a single EGT, QuatE \cite{quate2019zhang}, MuRE, ATTE, and 5*E \cite{nayyeri20215} unify at least two EGTs in their frameworks. 
The unification is achieved in QuatE and 5*E through Hamilton product and M\"obius transformation respectively. 
In contrast, ATTE's framework uses an attention mechanism to combine rotation and reflection transformations. 
Relations learn one or a mixture of the two transformations. 
On the other hand, MuRE integrates translation with matrix multiplication to transform head and tail entities simultaneously. 

\paragraph{\textbf{Ensemble approaches to knowledge graph embeddings.}}
Ensemble KGE models aggregate predictions from multiple embedding methods, aiming to outperform individual models. 
The concept was first introduced in \cite{krompass2015ensemble} where three base models are trained independently, and their triple plausibility predictions averaged. Later, the work \cite{chang2020benchmark}, extended this ensemble model to combine ten baseline models. 
More recent work by \citeauthor{xu2021multiple}~\cite{xu2021multiple} proposed averaging predictions from multiple parallel copies of a single KGE model trained in lower dimensions, summing to the original dimension. 
Another ensemble model \cite{gregucci2023link} combines prediction from SOTA models in two different ways: spherical and Riemannian attention-based query embeddings. In both cases, the attention is evaluated with a parametrised function which depends on the model prediction. Given a query, this mechanics allows the ensemble model to pay attention to the best performing KGE model. 

Notably, none of these ensemble approaches constrain relations to selectively prioritize or vote for specific top-performing geometric transformations. Instead, they focus on combining model outputs broadly or optimizing over complex transformation combinations.
In contrast, the framework we propose initializes each relation with all the available geometric transformations (EGTs) embeddings. It then trains the embeddings such that every relation ranks the EGTs based on its affinity to them. This allows the framework to single out the top relation-specific transformations per relation and return an optimal model based on different final selections such: embedding all relations with the most used EGT; embedding each relation with its top ranked EGT, or its first top ranked EGTs whose ranks are higher than a fixed threshold. 
Our goal is not to combine the strength of the EGTs in a single relation, as it is the case for the existing models, but to leverage the underlying KG structure for an optimal 
relation-EGTs association. 

\section{Preliminaries}\label{sec:preliminaries}
\subsection{Knowledge graph}
A Knowledge Graph (KG) is a multi-relational directed graph $ \mathcal{KG} =(\mathcal{E},$ $\mathcal{R},$ $\mathcal{F})$, where $\mathcal{E},$ $\mathcal{R}$ and $\mathcal{F}$ are the set of nodes (or entities), set of edges (or relations between entities), and a subset of triples (or facts) in $\mathcal{E}\times \mathcal{R}\times \mathcal{E}$, respectively. 
A Knowledge Graph Embedding (KGE) model learns vector representations of entities ($\mathcal{E}$) and relations ($\mathcal{R}$). 
In this paper, we embed entities and relations in a $d$-dimensional complex space. Entities in a KG are represented as $\mbf{e}, \mbf{h}, \mbf{t} \in \mbb{C}^{d},$ and relations as $\mbf{r}  \in \mbb{C}^{d}.$ We denote by $\tau$ the EGTs 
and use $\mbf{r}[\tau]$ to stress on the pairing relation-EGT embedding. 

\subsection{Elementary geometric transformations}
Elementary geometric transformations (EGTs) refer to fundamental operations such as translation, rotation, reflection, and scaling. The broader concept of geometric transformations (GTs) encompasses both these elementary geometric transformations and their combined transformations. 
GTs serve to move, reshape or reorient the latent representation of sub-graphs within embedding spaces, or in hyperplanes. 
In this work, we discuss the EGTs in the context of knowledge graph embedding models.

\noindent \textbf{Translation transformation.} Translation is an EGT that moves 
entities or sub-graphs from one position into another along a specific direction in the embedding space or coordinate plane. 
This makes translations particularly well-suited for modeling hierarchical relations. 
Intranslation-based embeddings, a translation vector $\mbf{u}
\in \mbb{C}^d,$ is applied to transform the embedded head entity, $\mbf{h}$ as 
\begin{equation}
    r(\mbf{u}, \mbf{h}) = \mbf{h} \oplus \mbf{u},
\end{equation}
where $\oplus$ is the elementwise complex addition. 
Notably, translation is a bijective mapping which is both antisymmetric and commutative. 

\noindent \textbf{Rotation and reflection transformation.} 
Rotation is a geometric transformation that moves entities or sub-graphs around an axis defined by a (unit) vector, while reflection mirrors the entities across to a fixed line or hyperplane. 
Rotation is an isometry, similar to translation. 
It is antisymmetric, unless its angle is a multiple of $\pi.$  Composition of rotations is commutative. 
It is limited in preserving tree-like structures as after applying it finitely many times, it returns the entity into its initial position. 
Reflection is a non-commutative symmetric isometry which is also unable to preserve tree like structures. 
Matrix representation of transformations in the complex plane is widely used in the literature. While a complex expression of rotation, 
$
    Rot(z) = e^{i\theta} z \text{ for } z\in \mbb{C},
$ is often provided, a similar expression for reflection has never been used, at least to the best of our knowledge. We therefore derived its expression as follows. 
A two-dimension matrix representation of the reflection transformation by an angle $\phi$ is denoted and defined by 
\begin{equation}
\label{eq: matrix rep of Ref}
    \mathit{Ref}(z) = \left(\begin{smallmatrix}
        \cos(2\phi) & \sin (2\phi)\\
        \sin (2\phi) & -\cos (2\phi)
    \end{smallmatrix}\right)
    \left(\begin{smallmatrix}
        x\\
        y
    \end{smallmatrix}\right)
\end{equation}
where $x, y\in \mathbb{R}$ are the real and imaginary parts of $z.$ Equation~\eqref{eq: matrix rep of Ref} yields 
$
\mathit{Ref}(z) = 
        x\cos(2\phi) + y\sin(2\phi) + i(
        x\sin(2\phi)  -y\cos(2\phi)).
$ 
Assuming that $r\cos\alpha +i r\sin \alpha$ is the trigonometric form of $\mathbf{h},$ then 
$
\mathit{Ref}(z) $=$ r\cos(2\phi - \alpha) +ir\sin(2\phi - \alpha) $=$ r e^{i(2\phi - \alpha)}.
$ 
It follows that 
$
    \mathit{Ref}(z) = e^{2i\phi}\bar{z}.
$ 
Thus, rotation and reflection embeddings of relations use $d$ dimensional unit complex numbers $e^{i\mbf{\theta}}$ and $e^{i\mbf{\phi}}$ to map the head embeddings to  
\begin{equation*}
    r(\mbf{\theta}, \mbf{h}) = e^{i\mbf{\theta}}\odot \mbf{h}
\quad \text{ and } \quad 
    r(\mbf{\phi}, \mbf{h}) = e^{2i\mbf{\phi}}\odot \bar{\mbf{h}}
\end{equation*}
respectively.  
$\theta, \phi\in (-\pi, \pi]^{d};$ and $ \odot, \ominus$ and\ $\bar{\ }$\ denote elementwise complex product, subtraction, and
conjugate. 
$\theta$ and $\phi$ serves as rotation and reflection  angles for transforming the embedded head entity. 

\noindent \textbf{Scaling transformation.} Scaling is a geometric transformation that shrinks or enlarges the latent representation of sub-graphs by a constant scalar. 
Relations do also scale the embedded head by using a purely real complex vector, $\mbf{s}\in \mbb{R}^d \subset \mbb{C}^d.$ We denote the scaled head entity by 
\begin{equation*}
    r(\mbf{s}, \mbf{h})  = \mbf{s}\odot \mbf{h}.
\end{equation*}
Unlike the first three transformations, scaling is not an isometry; however, it is commutative and antisymmetric. 

\smallskip
In general, we will denote by $r(\tau, \mbf{h})$ any of the four aforesaid transformed heads by a relation $r.$ $\tau$ could stand for the embedding vectors: $\mbf{u}, \theta, \phi, \mbf{s};$ or for the EGTs
.

\section{The SMART framework}\label{sec:smart_kge}
\begin{figure}[h!]
    \centering
    \includegraphics[width=10cm]{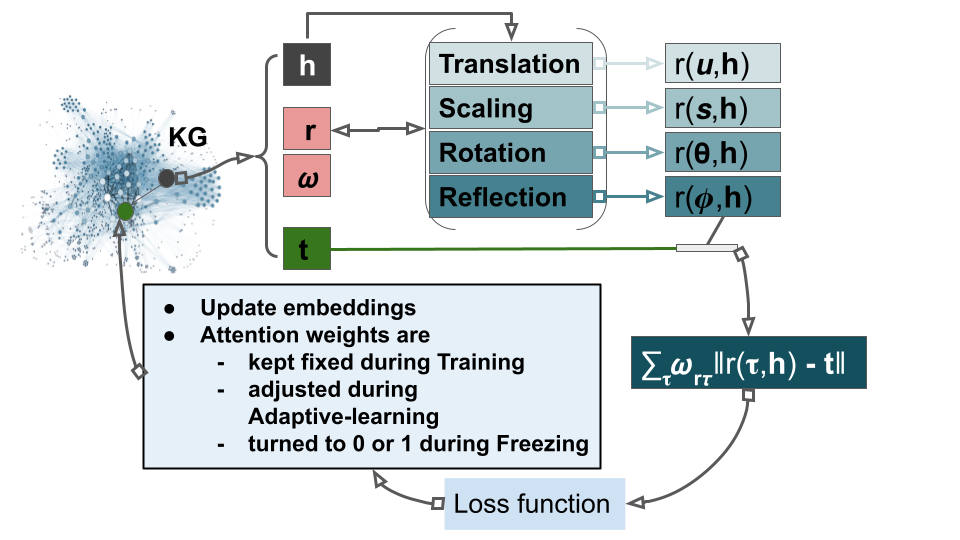}
    \caption{SMART framework. Three learning phases of SMART operated with relational attention weights in assigning EGTs to individual relations.}
    \label{fig: SMART Framework}
\end{figure}

SMART is a unified framework (illustrated in Figure~\ref{fig: SMART Framework}) built on a set of elementary EGTs, namely translation (\emph{Trans}), rotation (\emph{Rot}), and reflection (\emph{Ref}). The framework is modular and extensible, allowing for the incorporation of more complex GTs.
SMART operates through three sequential learning phases: training, adaptive learning, and freezing. These phases are coordinated with \textit{relation-specific attention weights} which guide the aggregation of the triple scores produced by the individual EGTs.


The full variant of the model, which undergoes three phases, is referred to as SMART. Simplified versions are denoted as SMART-X and SMART-XY, depending on which subset of phases (X or X and Y) they implement. 
In SMART-T, each relation is initialised with, and jointly trains, a set of four EGT embeddings.
In SMART-TA, attention weights are used to value the  contribution from each EGT per relation, enabling a relation-specific ranking of EGTs. SMART then proceeds by aggregating these rankings to determine the optimal transformation configuration for each relation, driven by the underlying structure of the KG. Finally, we define \sm{m},  a variant of SMART, which differs in how it aggregates the EGT rankings, employing one of two alternative strategies.

\subsection{Learning relational attention weights
\label{subsect: Attention mechanisms}} 
In the first step of learning, we uniformly initialise entity ($\mbf{e}$) and relation ($\mbf{u}, \theta, \phi, \mbf{s}$) embeddings which are jointly optimized during training. 
In addition to these embeddings, we introduce a \textit{relational attention weight matrix} $\mbf{W} = (\omega_{r{\tau}})_{r,{\tau}} \in [0, 1]^{n_r\times 4},$ where $n_r$ is the number of relations in the KG. 
$\mbf{W}$ is a learnable real-valued matrix whose row vectors sum to one; that is, $\sum_{\tau} \omega_{r{\tau}} = 1$ for all relations $r.$   
We initialize all the relational attention weights uniformly, assigning each $\omega_{r\tau}$ an initial equal value of $0.25,$ i.e $1$ divided by the total number of EGTs. These weights thereafter undergo three learning phases as follows.

\noindent \textbf{Training.} 
During this phase, all relational attention weights are kept fixed on $0.25.$ 
This ensures that all the four EGTs contribute equally to triple scores allowing the model to train each EGT uniformly and learn a diverse set of patterns. We refer to this phase as SMART-T which aims at reducing the impact of the randomness induced by the initialisation of entity and relation embeddings, providing a balanced starting point for further optimization.. 

\noindent \textbf{Adaptive-learning.} 
In this phase, the attention weights $\omega_{r\tau}$ are made trainable. The trained model (SMART-T) relaxes the constraint $\omega_{r\tau} = 0.25$ to $\sum_{\tau}\omega_{r\tau} = 1$ for all the relations. The weights are adapted according to the EGT-specific scores and turned to probability as 
\[
\omega_{r{\tau}}  = \frac{\exp\left(\omega_{r{\tau}}\right)}{\sum_{\tau'} \exp\left(\omega_{r{\tau'}}\right)}.
\]

Therefore, the higher the value of the weight $\omega_{r\tau},$ the stronger is the adherence of the relation $r$ to the EGT $\tau.$ 
This phase is an adaptive learning (SMART-TA) of the underlying structures resulting from mutual interaction between the relations.   
A persistently low performing EGT-based model decreases its corresponding weight gradually, since SMART-TA could only decrease the loss by decreasing its contribution.

\noindent \textbf{Freezing.} At the beginning of this phase, the highest 
relational attention weights obtained at the end of the adaptive learning phase are  set to $1$
, and $0$ otherwise. This allows the framework to assign the optimal EGT 
to each relation $r.$ 
Thus, low performing EGTs are systematically pruned
; only the remaining relation embeddings are optimized.  
The ultimate version  SMART-TAF is simply dubbed SMART.

\bigskip

The pruning process gives rise to two models. The main model, designated as SMART, involves the selection of EGTs based on the highest attention weights
\begin{equation}
\label{Eq: r-tau for SMART}
    r[\tau] =\mathrm{arg}\max_{\tau}\{\omega_{r\tau}\};
\end{equation}
$r[\tau]$ is the EGT assigned to the relation $r.$ 
The variant SMART\textsubscript{m} uses the majority voting to obtain $r_m[\tau]$ based on the $r[\tau]$s defined in Eq.~\eqref{Eq: r-tau for SMART}. That is  
\begin{equation}
\label{Eq: r-tau for SMART_m}
    r_m[\tau] =\mathrm{arg}\max_{\tau}\{\sum_{r}1 \left| r[\tau'] = \tau, \forall r \in \mathcal{R}\right.\}.
\end{equation}

In order to further mitigate the sensitivity of the framework in EGT assignment to the initialization of entity and relation embeddings, the model is retrained $n$ times, then 
\[
adh(r, \tau) = \sum_{j =1}^{n} \frac{\delta_{\{r[\tau_j] = \tau \}}}{n},
\]
 the average outcome per relation from these $n$ instances of optimal EGTs assigned by the model, is recorded; 
where $r[\tau_j]$ is the optimal EGT assigned to $r$ during the j'th run of SMART. $\delta_{\{r[\tau_i] = \tau \}}$ takes the value 1 if the assigned EGT is equal to $\tau,$ and 0 otherwise. 
Thus, $adh(r, \tau)$ quantifies the relational adherence.

\subsection{SMART plausibility function}
KGE performance depends directly on its ability to distinguish true triples from negative samples generated. 
The score function of some models such as DistMult and ComplEx is based on semantic similarity, while models such as TransE and RotatE use a distance-based score function. 
SMART belongs to the later KGEs category. 
The score of the triple $(h,r,t)$ is the weighted sum of the distance from the four transformed heads to the embedded tail entity
\[
 \Delta(h,r,t) = -\sum_{\tau} \omega_{r\tau}  \Vert r(\tau, \mbf{h}) - \mbf{t}\Vert.
\]
Since $\mbf{W}$ is a non-negative matrix, i.e. $w_{r\tau} \geq 0~ \forall (r, \tau) \in \mathcal{R}\times EGT,$ the score is negative. (We used $EGT$ to denote the set of EGTs.) 
True triples are expected to have high scores, whereas generated false triples gain small scores. 
For a given KG, triple scores guide SMART in learning the optimal embeddings of entities and EGTs. Furthermore, SMART increases (resp. decreases) the weight $\omega_{r\tau}$ if the score
of true triples with respect to the relation $r$ and its EGT embedding $\tau$ is high (resp. low). 
Otherwise, SMART pays more attention to the low-performing EGTs. 

\subsection{Formal properties}
Relations form diverse patterns as $\textit{premise} \rightarrow \textit{conclusion}$, where \textit{premise} can be a conjunction of several triples.
A KGE infers such relational patterns if the implication holds for the plausibility function. 
More precisely, if the score of all triples in the premise is positive, then the score for the conclusion must be positive as well.
SMART infers Symmetry/Antisymmetry, Inverse, Composition, and Commutativity/non-commutativity.

\noindent \textbf{Symmetry (Antisymmetry).} 
A relation $r \in \mathcal{R}$ is symmetric if both triples $(h, r, t)$ and $(t, r, h)$ coexist in the KG for all $h, t\in \mathcal{E};$ it is antisymmetric if it is never the case. 
SMART infers symmetry through its choice of reflection and rotation transformations; and antisymmetry through its choice of translation, rotation, and scaling transformations.

\noindent \textbf{Inversion.} The relation $r' \in $ $\mathcal{R}$ is the inverse of the relation $r$ if $(h, r, t) \rightarrow (t, r', h)$ holds when 
$ (h, r, t), (t, r', h) \in \mathcal{KG}.$ 
The four EGTs are invertible. More explicitly, the relation embeddings $(\mbf{u}, {\theta}, {\phi},\mbf{s})$ admit $(\mbf{-u}, {-\theta}, {\phi},\mbf{1/s})$ for inverse. 
Furthermore, the inverse of any finite composition of the EGTs is the composition of their inverses in the reverse order. 
SMART infers inversion. 

\noindent \textbf{Composition.} 
The relation relation $r_3$ is the composition of the relations $r_2$ and $r_1$ if  $(h, r_1, m) \land (m, r_2, t) \rightarrow (h, r_3, t)$ where $(h, r_1, m), (m, r_2, t), (h, r_3, t) \in \mathcal{KG}$.
It is clear that the composition of two different EGTs does not necessarily return another EGT. 
In the top sub-rows of Table~\ref{tab:com table}, we can see that the composition of two rotations 
yields a rotation (\textit{akin-composition}). 
In contrary, the composition of a rotation and a reflection is a reflection (\textit{unakin-composition}).  
SMART infers akin-composition by reducing the EGTs to a unique transformation similar to RotatE and TransE; otherwise it infers unakin-composition
. 

\noindent \textbf{Commutativity (non-commutativity).} 
Two relations $r_1 , r_2  \in $ $\mathcal{R}$ are said to be commutative, if $(h, r_1, m) \land (m, r_2, t)\rightarrow  (h, r_2, m) \land (m, r_1, t)$ where $(h, r_1, m),$ $ (m, r_2, t),$ $(h, r_2, m),$ $(m, r_1, t) \in \mathcal{KG}$.
Respectively, $r_1 , r_2$ are non-commutative relation if
$(h,$ $r_1,$ $ m),$ $(m, r_2, t)\in \mathcal{KG}$ but $(h, r_2, m) \notin \mathcal{KG} \lor (m, r_1, t)\notin \mathcal{KG}$.
The truth table of the propositional expression: ``composition of two EGTs is commutative", is shown in Table~\ref{tab:com table}. 
This shows the ability of SMART to infer commutative and non-commutative relations, by favouring scaling since it commutes with all EGTs but Translation. 
As a consequence, relational adherence to Translation is very negligible.

\section{Experiments}\label{sec:exp}

In this section we evaluate SMART against existing KGE models on four different datasets. 

\subsection{Evaluation protocol}  
We evaluate each KGE model on a KG Completion (KGC) downstream task, modeled as link prediction. 
The goal of link prediction is to predict the missing entity $?$ in the queries $(h, r, ?)$ or $ (?, r, t)$.
The missing entity is replaced by all entities in the KG to obtain corrupted triples. We filter the set of corrupted triples following TransE model~\cite{bordes2013transe}. 
SMART scores these triples and ranks them in a decreasing order. 
The predicted missing entity is 
the top ranked corrupted entity. 
To compare our models against the baselines and the EGT-based models, we used two metrics: the mean reciprocal rank (MRR), defined as $\sum_{j=1}^{n_t} \frac{1}{r_j}$ where $r_j$ is the rank of the $j$-th (positive) test triple and $n_t$ is the number of triples in the test set, and H@N which is the percentage of the triples whose rank is equal or smaller than $ N$ ($N=1,3, 10$).

\subsection{Datasets and benchmark models}

In our experiments we use the following four datasets: WN18RR \cite{dettmers2018convolutional}, FB15k-237 \cite{toutanova2015observed}, YouTube \cite{cen2019representation} 
and Company Ownership (CO) KG. 
The COKG is an anonymised sub-graph of the KG maintained by a prominent European central bank, where the nodes denote individuals or companies, and the edges illustrate shareholding relationships, such as ownership and control. The primary KG consists of about 1,657 million weakly connected components, and its degree distribution adheres to a scale-free power-law, which is typical in corporate economics scenarios. 
Table~\ref{tab:KGs Stats} summarizes the statistical information of the datasets. 
In terms of benchmark models, we compare SMART against the following EGT-based models, namely TransE (using translation),  DistMult (using scaling), RotatE (using rotation); and GT-based models, namely ComplEx (composition of scaling and rotation),  ROTE (composition of translation and rotation), REFE (composition of translation and reflection), ATTE (composition of translation and rotation or reflection), and MuRE (composition of translation and scaling).

\subsection{Hyperparameter search}
SMART 
is developed in the Rotat3D framework \cite{gao2020rotate3d}. 
We conducted an extensive hyperparameters grid search of SMART on the selected datasets and the optimal hyperparameter set was fixed based on the best MRR. 
Dimension $d=32$ is used for low dimension embedding, and  for high dimension embedding we fixed $d=100$ for COKG, $d=250$ for WN18RR, and $d=300$ for YouTube. 
The batch size, $\beta,$ and the number of negative sample, $\eta,$ are tuned in the range of $\{128, 256, 512, 1024\};$ the margin $\gamma$ is in $\{1, 9, 24\},$ the self-adversarial temperature $\alpha$ in $\{0, 0.5, 1\},$ the learning rate $\lambda$ in $\{0.0001, 0.001\};$ and the regularizer $\rho$ in $\{0,$ $ 0.01,$ $ 0.0001\}.$ 
The maximum step size $\sigma$ is set to $120000$ for the Training phase, $50000$ for the Adaptive-learning phase, and $90000$ for the Freezing phase. We emplemented an early stop mechanism to stop the learning in each phase when the MRR starts decreasing. This mechanism can be applied across the phases to allow the framework to prioritise between SMART-T, SMART-TA and SMART. The Adam optimizer was used as the optimization function. 
The optimal hyperparameter set is used to train and evaluate SMART and \sm{m} 20 rounds of run in order to compute the mean and standard deviation of the evaluation metrics.

\begin{table}
\small
    \centering
    \caption{Result of composing two elementary geometric transformations and the logical validation of their commutativity. ``$\notin EGT$" means the result is not an EGT.}
    \begin{tabular}{lrrrr}
    \hline
        $\circ $ & Translation & Rotation & Reflection & Scaling \\
        \hline
        \multirow{2}{*}{Translation}& Translation & $\notin EGT$& $\notin EGT$& $\notin EGT$\\
         & True & False &  False & False\\
        \hline
        \multirow{2}{*}{Rotation}& $\notin EGT$ & Rotation & Reflection & $\notin EGT$\\
         & False & True & False & True\\
        \hline
        \multirow{2}{*}{Reflection}  & $\notin EGT$ & Reflection & Rotation & $\notin EGT$\\
        & False & False & False & True \\
        \hline
        \multirow{2}{*}{Scaling} & $\notin EGT$ & $\notin EGT$ & $\notin EGT$ & Scaling\\
        & False & True & True & True \\
        \hline
    \end{tabular}
    \label{tab:com table}
\end{table}

\begin{table}[h]
\small
      \centering
      \caption{Dataset Statistics. 
      }
    \label{tab:KGs Stats}
    \begin{tabular}{lrrrrr}
         \toprule
         Dataset & \#E & \#R & \#Train &\#Valid. &\#Test  \\
         \midrule
         WN18RR & $40943$ & $11$ & $86835$ & $3034$  & 3134\\
         FB15k-237 & $14541$ & $237$ & $272115$ & $17535$  & 20466\\
         YouTube & $2000$ & $5$ & $1114025$ & $65512$ & $131007$\\
         COKG & $29268$ & $7$ & $57285$ & $6365$ & $7073$ \\
         \bottomrule 
    \end{tabular}
\end{table}

\begin{table*}[ht!]	
\centering 
\small

\caption{Results on WN18RR and FB15k237 in dimension 32. 
-- 
 $^a$ are reported from \cite{chami2020low}}
		\begin{tabular}{lllllllll}
                \toprule 
		& \multicolumn{4}{c}{\textbf{WN18RR}} & \multicolumn{4}{c}{\textbf{FB15k-237}} \\
			\cmidrule(lr){2-5}                  
			\cmidrule(lr){6-9}
			 & \textbf{MRR} & \textbf{H@1} & \textbf{H@3} & \textbf{H@10} 
              & \textbf{MRR} & \textbf{H@1} & \textbf{H@3} & \textbf{H@10} \\\midrule
    TransE & $366$ &  $27.4$ & $43.3$ & $51.5$ & $295$ &  $21.0$ & $32.2$ & $46.6$\\
    RotatE$^a$ & $387$ &  $33.0$ & $41.7$ & $49.1$ & $290$ &  $20.8$ & $31.6$ & $45.8$\\	
    ComplEx$^a$ &  $420$ & $39.0$ & $42.0$ & $46.0$ & $294$ & $21.1$ & $32.2$ & $46.3$\\	
    MuRE$^a$ &  $458 $ & $ 42.1$ & $47.1$ & $52.5$ &  $313 $ & $22.6$ & $34.0$ & $48.9$\\	
    REFE$^a$&  $455$ & $41.9$ & $47.0$ & $52.1$ &  $302$ & $21.6$ & $33.0$ & $47.4$\\	
    ROTE$^a$&  $463 $ & $42.6$ & $47.7$ & $52.9$ &  $307$ & $22.0$ & $33.7$ & $48.2$\\	
    ATTE$^a$& $456$ & $41.9$ & $47.1$ & $52.6 $ & $311$ & $22.3$ & $33.9$ & $48.8 $\\	
    \midrule 
    SMART & $449\pm 2$ & $40.8\pm 0.3$ & $46.0\pm 0.2$ & $53.4\pm 0.3$  & $274\pm 4$ & $18.4\pm 0.4$ & $30.4\pm 0.4$ & $45.6\pm 0.3$ \\
     \sm{m} & $438\pm 11$ & $39.7\pm 0.6$ & $45.1\pm 1.7$ & $52.2\pm 2.0$  & $298\pm 1$ & $20.5\pm 0.1$ & $33.0\pm 0.1$ & $48.7\pm 0.1$ \\
			\bottomrule
		\end{tabular}  
	\label{tb:fbwnresults}
\end{table*}
\begin{table*}[ht!]
    \centering \small
    \caption{Link prediction results on YouTube in high (300) and low (32) dimensions. We evaluated the baseline models
    .}
    \begin{tabular}{clllllllll}
    \toprule 
    & \multicolumn{4}{c}{High dimension} && \multicolumn{4}{c}{Low dimension}\\
    \cline{2-5} \cline{7-10}
    Model	& MRR &  H@1 & H@3 & H@10	&& MRR &  H@1 & H@3 & H@10\\	
    \midrule 
    TransE &  $180$ & $0$ & $29$ & $47$  
     &&  $226$ & $15$ & $24$ & $37$\\	
    RotatE &  $250$ &  $14$ & $30$ & $47$ 
     &&  $233$ & $12$ & $28$ & $45$\\	
    ComplEx &  $320$ & $21$ & $36$ & $54$ 
     &&  $306$ & $20$ & $35$ & $52$\\	
    DistMult & $37$ & $2$ & $3$ & $7$ 
     && $37$ & $2$ & $3$ & $7 $\\	
    SMART & $294\pm 3 $ & $15\pm 0$ & $38\pm 0$ & $55\pm 0$
    && $246\pm 7$ & $14\pm 1$ & $28\pm 1$ & $45\pm 1$\\     
    \sm{m} & $296\pm 1$ & $17\pm 0$ & $36\pm 0$ & $52\pm 0$
     && $247\pm 11$ & $12\pm 2$ & $30\pm 1$ & $49\pm 1$\\
    \bottomrule
    \end{tabular}
        \label{tab:YT dim 32}
\end{table*}
\begin{table*}[ht!]
    \centering \small
    \caption{Link prediction on WN18RR with (denoted in line for True) and without (False) loading adherence embeddings. 
    }
    \begin{tabular}{ccccccccccccc}
    \toprule
    & Load adh.& \multicolumn{4}{c}{Evaluation} & & \multicolumn{6}{c}{Best Configuration}\\
    \cmidrule(lr){3-6} \cmidrule(lr){7-13}
    Model & embeddings? & MRR &  H@1 & H@3 & H@10 & &$\gamma$ & $\beta$ & $\alpha$ & $\eta$ & $\lambda$  & $\rho$\\
    \midrule
    \multirow{2}{*}{\sm{} }& \multirow{1}{*}{True} &  $471$ & $43.0$ & $48.6$ & $55.1$ & & $9$ & $1024$ & $0$ & $512$ & $0.0001$ & $0.1$\\
    & \multirow{1}{*}{False} & $467$ & $42.5$ & $48.2$ & $55.4$  & & $9$ & $1024$ & $0$ & $512$ & $0.0001$ & $0.1$\\
    \multirow{2}{*}{\sm{m} }& \multirow{1}{*}{True} &  $453$ & $41.3$ & $46.8$ & $53.3$ & & $24$ & $256$ & $0$ & $512$ & $0.0001$ & $0.1$\\
    & \multirow{1}{*}{False} & $454$ & $41.1$ & $46.6$ & $54.5$  & & $9$ & $512$ & $1$ & $512$ & $0.0001$ & $0$\\
    \bottomrule
    \end{tabular}
    \label{tab: Adherence transfer}
\end{table*}

\subsection{Findings and directions for improvement}
\noindent \textbf{Evaluation results in low dimension.}
We discuss in this section the empirical results of SMART and its variants for the KGC task in \emph{low dimension}. 
We reported in Tables \ref{tb:fbwnresults} - \ref{tab:COKG 32} the mean value and the standard deviation of SMART performances after multiple runs. However, we will only consider the mean values while comparing the models. This randomness comes from initializing the relation attention weights and the entity and relation embeddings. 
The results in Table~\ref{tb:fbwnresults} show that \sm{} and \sm{m} surpass the EGT-based models on WN18RR and FB15k-237. On the other hand, the GT-based models, except ComplEx, outperform our models on MRR, H@1, and H@3. This demonstrates that composite transformations are more convenient in modeling relations in these two KGs. Our models may also benefit from substituting the EGTs with the GTs. 
We observed \sm{m} is the second-best performing model, behind ComplEx, on YouTube
. RotatE and \sm{} perform similarly on COKG, yet TransE appears to gain more from using only Translation. Although \sm{} could degrade to, or \sm{m} could enforce TransE,  the architecture of the framework prevented them from doing so. As it can be seen from the weights distribution in Table~\ref{tab:Adherence PTF 1259}, none of the relations adhered to translation (results in \emph{standar black values}). The framework could therefore benefit from a more intelligent mechanism. 
In general, the results show that our models are comparable to the state of the art, with the advantage of being able to select relation-specific GTs.
\begin{table*}[ht!]
\centering \small
    \caption{Link prediction results on COKG in high (100) and low (32) dimensions. We evaluated baseline models on COKG.
    }
    \begin{tabular}{clllllllll}
    \toprule 
    & \multicolumn{4}{c}{High dimension} && \multicolumn{4}{c}{Low dimension}\\
    \cline{2-5} \cline{7-10}
    Model	& MRR &  H@1 & H@3 & H@10 && MRR &  H@1 & H@3 & H@10\\	
    \cline{1-5} \cline{7-10}
    TransE &  $428$ & $34.9$ & $47.5$ & $56.4$ &&  
    $552$ & $52.1$ & $56.4$ & $61.7$\\	
    RotatE &  $489$ & $42.0$ & $54.2$ & $58.9$ &&
     $548$ & $52.5$ & $55.9$ & $59.6$\\	
    ComplEx &  $154$ & $11.5$ & $13.3$ & $24.9$ &&
     $517$ & $46.7$ & $55.0$ & $61.9$\\	
    DistMult & $388$ & $31.3$ & $43.9$ & $55.5$ &&
    $439$ & $39.4$ & $45.2$ & $54.8$\\	
    SMART & $550\pm 4$ & $50.9\pm 0.6 $ & $58.9\pm 0.2$ & $61.4\pm 0.1$&& 
    $542\pm 13$ & $50.8\pm 1.7$ & $56.9\pm 1.6$ & $60.2\pm 0.6$\\
    \sm{m} & $421\pm 10$ & $37.5\pm 1.7$ & $44.6\pm 0.7$ & $50.6\pm 0.5$&&
    $496\pm 40$ & $45.9\pm 4.2$ & $52.5\pm 4.5$ & $55.4\pm 3.9$\\
    \bottomrule
    \end{tabular}
        \label{tab:COKG 32}
\end{table*}

\begin{table*}[h!]
\small
    \centering 
\caption{Relational adherence (in \%) to the EGTs 
in low dimension.
The \emph{standard black weights}  correspond to the default EGT order: \emph{Trans, Rot, Ref, Scal}; and the \textcolor{blue}{\emph{italicized blue weights}}  correspond to the alternative order: \emph{Trans, Scal, Ref, Rot}.}
    \begin{tabular}{lcccc|lc|cc|cc|cc|c}
   \toprule 
   &\multicolumn{4}{c}{WN18RR} &   \multicolumn{9}{c}{COKG}\\
    \midrule
      Relation &\emph{Translation} & \emph{Rotation} & \emph{Reflection} & \emph{Scaling} 
      & \multicolumn{1}{c}{Rel.} &\multicolumn{2}{c}{\emph{Translation}} & \multicolumn{2}{c}{\emph{Rotation}} & \multicolumn{2}{c}{\emph{Reflection}} & \multicolumn{2}{c}{\emph{Scaling}}\\
     \midrule 
\emph{hasPart} & $0$  & $\bf 100$  & $0$  & $0$ 
& \emph{Rel.A}  & $0$ & $\tc{0}$  & ${\bf 50}$ & $\tc{36}$  & $\bf 50$ & $\tc{\bf 42}$  & $0$ & $\tc{21}$\\
\emph{\_meronym} & $0$  & $\bf 99$  & $1$  & $0$ 
& \emph{Rel.B} & $ 0$ & $\tc{0}$  & $\bf 50$ & $\tc{33}$  & $\bf 50$ & $\tc{\bf 39} $ & $0$ & $\tc{28}$\\
\emph{region} & $0$  & $\bf 90$  & $4$  & $6$ 
&\emph{Rel.C} &  $0$ & $\tc{0}$  & $\bf 52$ & $\tc{29}$  & $48$ & $\tc{30}$  & $0$ & $\tc{\bf 41}$\\
\emph{hypernym} &$ 0$  &$\bf 100$  &$ 0 $  &$ 0$ 
& \emph{Rel.D} & $0$ & $\tc{0}$  & $\bf 56$ & $\tc{36}$  & $42$ & $\tc{22}$  & $2$ & $\tc{\bf 42}$\\
\emph{topicOf} & $0$  & $\bf 42$  & $40$  & $20$ 
& \emph{Rel.E} & $1$ & $\tc{2}$  & $26$ & $\tc{19}$  & $26$ & $\tc{19}$  & $\bf 47$ & $\tc{\bf 61}$\\
\emph{alsoSee} & $0$  & $0$  & $\bf 100$  & $0$ 
&  \emph{Rel.F} & $0$ & $\tc{0}$  & $\bf 51$ & $\tc{33}$  & $49$ & $\tc{30}$  & $0$ & $\tc{\bf 37}$ \\
\emph{similaTo} & $0$  & $10$  & $\bf 90$  & $0$ 
&  \emph{Rel.H} & $1$ & $\tc{19}$ & $\bf 38$ & $\tc{\bf 28}$ & $\bf 38$ & $\tc{28}$ & $14$ & $\tc{25}$\\
\cline{6-14} 
\emph{verb group} & $0$  & $2$  & $\bf 98$  & $0$ 
&  \multicolumn{9}{c}{YouTube}\\
\cline{6-14}
\emph{ins.\_hypernym} & $0$  & $20$  & $\bf 80$  & $0$ 
& \emph{Rel.3} & $30$ & $\tc{26}$  & $\bf 50$ & $\tc{\bf 51}$  & $17$ & $\tc{18}$  & $3$ & $\tc{4}$\\
\emph{\_usage} & $0$  & $9$  & $2$  & $\bf 74$ 
& \emph{Rel.4} & $26$ & $\tc{25}$  & $\bf 44$ & $\tc{\bf 45}$  & $18$ & $\tc{19}$  & $13$ & $\tc{12}$\\
\emph{deriv.Related} & $0$  & $\bf 100$  & $0$  & $0$ 
& \emph{Rel.5}  & $31$ & $\tc{27}$  & $\bf 36$ & $\tc{\bf 37}$  & $17$ & $\tc{18}$  & $16$ & $\tc{19}$\\
    \bottomrule
     \end{tabular}
    \label{tab:Adherence PTF 1259}
\end{table*}

\noindent \textbf{Transferring EGT preferences across 
dimensions.} We evaluated both the baseline and our models in dimension 300 on the YouTube dataset. As shown in Table~\ref{tab:YT dim 32}, increasing the embedding dimension generally improved the performance of all models, except for TransE, whose MRR and H@1 considerably declined.

\begin{table*}[!ht]	
\centering 
\caption{Optimization across successive phases. $\mu$ and $\sigma,$ 
stand for mean and standard deviation 
of the performances after multiple runs.
}
		\begin{tabular}{lllllllllll}
                \toprule 
		& & \multicolumn{3}{c}{\textbf{WN18RR}} & \multicolumn{3}{c}{\textbf{FB15k-237}}& \multicolumn{3}{c}{\textbf{YouTube}} \\
			\cmidrule(lr){3-5}                  
			\cmidrule(lr){6-8}
            \cmidrule(lr){9-11}
		\textbf{Model}	&  \textbf{Stats} & \textbf{MRR} & \textbf{H@1}  & \textbf{H@10} 
              & \textbf{MRR} & \textbf{H@1}
              & \textbf{H@10} 
              & \textbf{MRR} & \textbf{H@1}
              & \textbf{H@10}\\\midrule
\multirow{1}*{SMART-T}
&$\mu \pm \sigma$ &$ 397\pm 7$&$ 34.4\pm 1.3 $&$ 50.5\pm 0.2 $&${\bf 279}\pm 1$&$ {\bf 19.0}\pm 0.2 $&$ \underline{ 46.0}\pm 0.1$ &$215\pm 3$&  $9.9\pm 0.6$&  ${43.7}\pm	0.2$\\
\multirow{1}*{SMART-TA}
&$\mu \pm \sigma$ & $\underline{446}\pm 2$ & $\underline{40.4}\pm 0.2$ & $\underline{53.1}\pm 0.3$ & ${\bf 279}\pm 1$ & $\underline{18.9}\pm 0.1$ & ${\bf 46.2}\pm 0.2$ &$\underline{231}\pm 5$ &  $\underline{11.7}\pm 0.8$ & $\underline{45.0}\pm 0.5$\\
\multirow{1}*{SMART}
&$\mu \pm \sigma$	& ${\bf 449}\pm 2$ & ${\bf 40.8}\pm 0.3$ & ${\bf 53.4}\pm 0.3$ & $\underline{274}\pm 4$ & $18.4\pm 0.4$ & ${45.6}\pm 0.3$ &${\bf 246}\pm 7$ &  ${\bf 14.2}\pm	0.7$ & ${\bf 45.1}\pm 1.1$ \\
			\bottomrule
		\end{tabular}  
	\label{tab:ablation study Influence}
\end{table*}

The performance gap between ComplEx and \sm{} narrowed, with \sm{} achieving a higher H@10 score than ComplEx. 
Embedding KGs in high-dimensional spaces allows the models to preserve the underlying subgraph structures more effectively. 
However, the size of the KGs, particularly the primary Company Ownership KG, makes extensive hyperparameter search in high dimensions computationally challenging. Since learned relational properties should be preserved across embedding spaces, regardless of dimensionality, we hypothesized 
that adherence learned in a low dimension could be reused in a high dimension without a high loss in performance. 
For the experiment, we performed the grid search in dimension 32, did multiple runs of SMART, and saved the relational adherence to EGTs. 
Without the training and adaptive learning phases, we fine-tuned a version of our models on WN18RR with the learned relational adherence in dimension $250$. 
A second version underwent the three-phase training procedure. In both scenarios,  we performed an extensive hyperparameter search. 
The results are in Table~\ref{tab: Adherence transfer}, where \emph{True} indicates that the relational adherence embeddings were loaded, and \emph{False} indicates they were not. Not only is the adherence embeddings loading computationally less expensive and comparable to the trained model (case of \sm{m}), but it can also be more accurate (case of \sm{}). 
We push the experiment further by training in dimension 100, all the models with their corresponding optimal hyperparameters learned in dimension 32 on the COKG. Table~\ref{tab:COKG 32} further supports our hypothesis. 
Additionally, we observed that reusing optimal hyperparameters leads to a decrease in performance for the baseline models, including \sm{m}. In contrast, it benefits \sm{}, enabling it to outperform all other models.

\noindent \textbf{Adaptive early stopping
.}
We conducted a study on implementing an early stopping mechanism across the three phases of the SMART framework. 
We computed the 
mean ($\mu$) and standard deviation ($\sigma$) statistics from multiple runs of SMART on WN18RR, FB15k-237, and YouTube (see Table~\ref{tab:ablation study Influence}). 
On WN18RR and YouTube,  we observed a consistent improvement in performance across phases. Consequently, \sm{}-T underperforms compared to \sm{}-TA, which in turn underperforms compared to the full \sm{} model. 
Results on FB15k-237 show that \sm{}-T and \sm{}-TA achieve comparable performance, both outperforming SMART. This suggests that the model can benefit from an early termination of the adaptive phase, while avoiding the freezing phase, which appears to impact final performance negatively. 
These results highlight the practical benefit of incorporating an adaptive early-stopping strategy, allowing the framework to select the best-performing phase dynamically rather than always proceeding through the full pipeline. We conclude that while all three phases are valuable, early termination at the optimal phase can yield better generalization and training efficiency.

\noindent \textbf{Empirical study on the ordering effect of EGTs.}
By default, the column vectors of the attention weight matrix are mapped to the EGTs in the following default order: \emph{Trans}, \emph{Rot}, \emph{Ref}, and \emph{Scal}. This fixed ordering introduces a bias in \sm{}, favoring earlier EGTs, e.g., Translation and Rotation. 
For instance, \emph{Rel.B} and \emph{Rel.H} are both assigned to Rotation
, even though Reflection 
holds an equal attention weight, as shown by the \emph{standard black values} in Table~\ref{tab:Adherence PTF 1259}. To investigate the impact of this ordering bias, we reassigned the EGTs to the column vectors in a new order: 
\emph{Trans, Scal, Ref, Rot}
, and reported the results in blue on the right side of the default values. 
While the adherence to EGTs in most KGs shows no sensitivity to this change, although the weights have slightly changed, it is notably affected in COKG. \emph{Rel.B} shifts significantly, now favoring \emph{Ref}, \emph{Rel.C} to \emph{.F} favor Scaling
, whereas \emph{Rel.H} continues to assign equal weights to Rotation and Reflection
. This observation highlights how EGT order can influence relational preferences and, consequently, model behavior.

\noindent \textbf{EGT selection using a confidence threshold.}
Driven by the possibility of obtaining equal attention weights and selecting only the initially encountered EGT, we chose to investigate the effect of allocating multiple EGTs to the relations. Consequently, we developed the final variant, SMART\textsubscript{>}, which opts for the top-performing EGTs by applying a threshold to the attention weights. 
Given  a threshold $\varepsilon,$ the selection is  
$
    r[\tau_*] =\{\tau \mid \omega_{r\tau} > \varepsilon\}.
$ 
We note that $\tau_*$ is a relation-specific subset of EGTs. From Table~\ref{tab:Adherence PTF 1259}, the weights displayed in black illustrate that $\tau_* =$\{\emph{Rot,~Ref}\} is assigned to all the relations in COKG except \emph{Rel.E} when $\varepsilon = 35\%$. 
In general, $\tau_*$ can be the empty set  if $\varepsilon$ is too high, or contains all the EGTs if $\varepsilon$ is too low. 
\begin{table}[!ht]	
\centering 
\caption{Performances of \sm{>} on COKG while loading adherence embeddings reported in Table~\ref{tab:Adherence PTF 1259}.}
		\begin{tabular}{lllll}
                \toprule 
		model &\textbf{$\varepsilon$}	& \textbf{MRR} & \textbf{H@1}   &  \textbf{H@10} \\
\midrule
\multirow{2}*{\sm{>}} &$0.25 $&$ 463\pm 7$&$ 41.9\pm 0.3 $&$ 54.3\pm 1.7$\\
& 0.35 & $527\pm 10$&$ 49.1\pm 1.6$&$ 59.5\pm 0.3$\\
\sm{} & ---&$542\pm 13$ & $50.8\pm 1.7$ & $60.2\pm 0.6$\\ 
			\bottomrule
		\end{tabular}  
	\label{tab:threshold}
\end{table}
Table~\ref{tab:threshold} demonstrates that opting for a single EGT appears to be more suitable than choosing multiple EGTs for KGC on the COKG dataset.

\section{Conclusion}\label{sec:conclusions}
We presented SMART -- a novel and unified framework for optimising the use of elementary geometric transformations for KGEs. The main novelty is that it allows a relation-specific selection of an appropriate transformation. We analysed the formal properties of the approach and showed that it covers various relation patterns. 
SMART is designed with three learning phases, and each of them as well as their combination, is analysed with regard to elementary geometric transformations they assign to relations. We proved that the framework benefits from an adaptive early stopping on the optimal phase. 
Through our assessment, we showed the generalization capabilities of SMART, which facilitate the transfer of EGT preferences and optimal hyperparameter setups across different dimensions, contributing to decreased computation time.  
To the best of our knowledge, SMART is the first KGE framework that investigates individual geometric transformations in connection to relations and their structural and relational patterns. 
SMART can be extended beyond elementary geometric transformations, such as composite transformations.

\begin{ack}
We acknowledge the financial support by the Federal Ministry of Education and Research of Germany and by S\"achsisches Staatsministerium f\"ur Wissenschaft, Kultur und Tourismus in the programme Center of Excellence for AI-research „Center for Scalable Data Analytics and Artificial Intelligence Dresden/Leipzig" (ScaDS.AI).
The work is also supported by the SECAI project (grant 57616814) funded by DAAD (German Academic Exchange Service), as well as the Leipzig University for providing GPU servers part of the evaluations. 
\end{ack}

\bibliography{mybibfile}

\end{document}